\documentclass[journal]{IEEEtran} 
\IEEEoverridecommandlockouts
\usepackage{cite}
\usepackage{amsmath,amssymb,amsfonts}
\usepackage{algorithmic}
\usepackage{algorithm}
\usepackage{graphicx}
\usepackage{textcomp}
\usepackage{xcolor}
\usepackage{booktabs}
\usepackage{multirow}
\usepackage{multicol}
\def\BibTeX{{\rm B\kern-.05em{\sc i\kern-.025em b}\kern-.08em
    T\kern-.1667em\lower.7ex\hbox{E}\kern-.125emX}}
\begin{document}

\title{CLIP-aware Domain-Adaptive Super-Resolution}

\author{
\author{
	\IEEEauthorblockN{Zhengyang Lu\textsuperscript{1}, Qian Xia\textsuperscript{2}, Weifan Wang\textsuperscript{1} and Feng Wang\textsuperscript{1*}}\\
	\IEEEauthorblockA{\textsuperscript{1} Jiangnan University, China\\
		\textsuperscript{2} Changshu Institute of Technology, China\\
		\thanks{Z.Lu: luzhengyang@jiangnan.edu.cn}}
}

}


\maketitle

\begin{abstract}
This work introduces CLIP-aware Domain-Adaptive Super-Resolution (CDASR), a novel framework that addresses the critical challenge of domain generalization in single image super-resolution. By leveraging the semantic capabilities of CLIP (Contrastive Language-Image Pre-training), CDASR achieves unprecedented performance across diverse domains and extreme scaling factors. The proposed method integrates CLIP-guided feature alignment mechanism with a meta-learning inspired few-shot adaptation strategy, enabling efficient knowledge transfer and rapid adaptation to target domains. A custom domain-adaptive module processes CLIP features alongside super-resolution features through a multi-stage transformation process, including CLIP feature processing, spatial feature generation, and feature fusion. This intricate process ensures effective incorporation of semantic information into the super-resolution pipeline. Additionally, CDASR employs a multi-component loss function that combines pixel-wise reconstruction, perceptual similarity, and semantic consistency. Extensive experiments on benchmark datasets demonstrate CDASR's superiority, particularly in challenging scenarios. On the Urban100 dataset at $\times$8 scaling, CDASR achieves a significant PSNR gain of 0.15dB over existing methods, with even larger improvements of up to 0.30dB observed at $\times$16 scaling.
\end{abstract}


\section{Introduction}

Single-image super-resolution (SISR) advances the development of image reconstruction techniques, as the frontier challenge in the low-level computer vision field \cite{wang2020deep,cui2023selective,cui2024omni}. 
SISR has become a critical component in fields as diverse as medical diagnostics, where it enhances the clarity of MRI scans \cite{chen2018efficient}, to space exploration, where it sharpens satellite imagery of distant planets \cite{lei2017super}. 
Deep learning has catapulted SISR into a new benchmark, with convolutional neural networks (CNNs) and transformers achieving unprecedented levels of detail recovery \cite{ledig2017photo,cui2024revitalizing,lu2024self}. 
These advancements have not only improved image quality metrics but have also facilitated facial recognition \cite{lu2021face}, forensic image analysis \cite{mayer2018learned}, and the preservation of cultural heritage through the restoration of historical photographs \cite{wan2020old,cui2023focal,lu2025single}. 
However, as resolution enhancement techniques approach theoretical limits, a fundamental tension emerges: the trade-off between computational efficiency and the quality of reconstructed images. 
The deeper exploration of super-resolution uncovers questions that challenge current understanding of image formation, perception, and the very nature of visual information itself \cite{lu2023joint, wang2021towards}.

\begin{figure}[t]
	\centering
	\includegraphics[width=\linewidth]{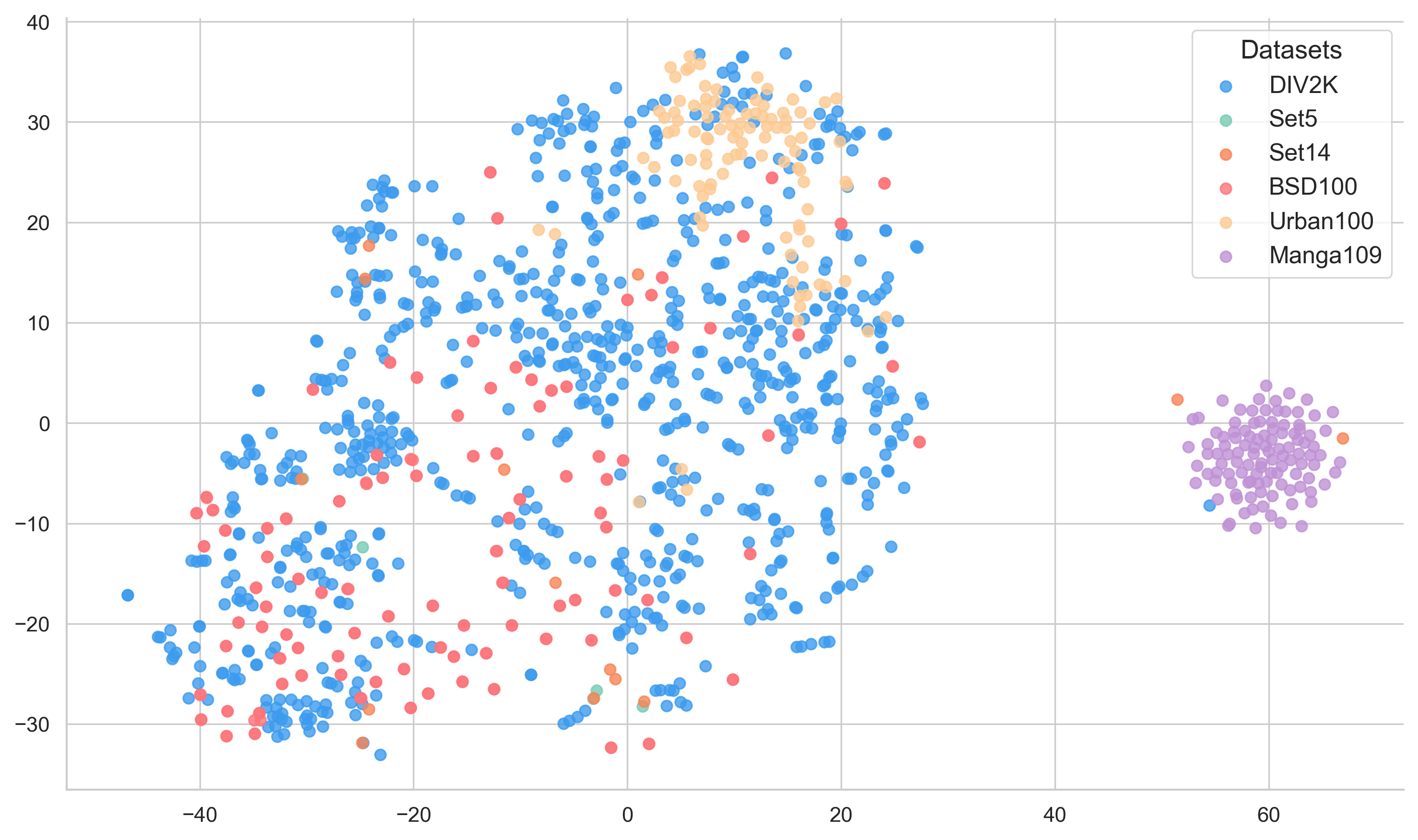}
	\caption{2D t-SNE visualization of CLIP features extracted from various super-resolution benchmark datasets. DIV2K (blue) shows a wide distribution covering multiple domains, BSD100 (red) clusters predominantly in one region with partial overlap with DIV2K, while Manga109 (purple) forms a distinctly separate cluster due to its unique anime-style content. This distribution highlights the significant domain differences that super-resolution models must address.}
	\label{fig:clip_features}
\end{figure}

Figure \ref{fig:clip_features} illustrates the 2D t-SNE visualization of CLIP features across various super-resolution datasets.  The distribution reveals interesting patterns in how these datasets relate in feature space.  DIV2K exhibits a wide distribution range, partially encompassing Urban100 features in one region, indicating some shared characteristics between these datasets.  However, Manga109 forms a distinctly separate cluster due to its unique anime-style content, which differs significantly in visual semantics from photographic images.  Even within the seemingly closer clusters, subtle separations are visible between dataset features, suggesting domain-specific characteristics that may impact super-resolution performance.  This visualization provides evidence that semantic feature distributions vary across datasets, with the degree of variation potentially influencing how well super-resolution models generalize across domains.  While some domains show greater similarity (DIV2K and Urban100), others represent more challenging transfer learning scenarios (Manga109), underscoring the need for domain-adaptive approaches in super-resolution tasks.

To address these challenges, we propose CLIP-aware Domain-Adaptive Super-Resolution (CDASR), a novel framework that leverages the semantic understanding capabilities of CLIP (Contrastive Language-Image Pre-training) to enhance domain adaptation in SISR. Our approach integrates a CLIP-guided feature alignment mechanism with a meta-learning inspired few-shot adaptation strategy, enabling efficient knowledge transfer across diverse image domains. The CLIP-guided feature alignment utilizes the rich semantic representations learned by pre-trained CLIP models to bridge the domain gap between source and target datasets. This alignment is achieved through a custom domain-adaptive module that processes CLIP features alongside super-resolution features, facilitating effective cross-domain knowledge transfer. The module employs a multi-stage transformation process, including CLIP feature processing through a multi-layer perceptron, spatial feature generation via convolutional layers, and feature fusion with the original super-resolution features. This intricate process ensures that the semantic information captured by CLIP is effectively incorporated into the super-resolution pipeline, allowing for generalizable reconstructions. Additionally, the framework incorporates a novel multi-component loss function that combines pixel-wise reconstruction, perceptual similarity, and semantic consistency, further enhancing the model's ability to preserve low-level details and high-level semantic information across diverse image domains.

Furthermore, we introduce a meta-learning inspired few-shot adaptation technique that allows our model to quickly adapt to target domains using only a few labelled examples. This approach leverages a meta-objective that optimizes for rapid adaptation across multiple domains, resulting in a flexible super-resolution model. The combination of CLIP-guided alignment and meta-learning enables CDASR to achieve state-of-the-art performance across multiple scaling factors, while requiring minimal fine-tuning for target domains.

Our experimental results demonstrate the effectiveness of CDASR across multiple benchmark datasets and scaling factors. On the challenging Urban100 dataset at $\times$8 scaling, CDASR achieves a PSNR gain of 0.15dB over the next best method, with even larger improvements of up to 0.30dB observed at $\times$16 scaling. These results highlight the robustness of our approach on extreme upscaling scenarios.

The main contributions of this work are summarized as follows:

\begin{itemize}
	\item We propose CDASR, a CLIP-aware domain-adaptive super-resolution framework that integrates CLIP's semantic representations with super-resolution features to enhance cross-domain generalization in SISR tasks.
	
	\item We introduce a CLIP-guided feature alignment mechanism with a multi-component loss function, ensuring both low-level fidelity and high-level semantic coherence across diverse domains.
	
	\item We develop a meta-learning inspired few-shot adaptation strategy, enabling rapid adaptation to new domains with minimal samples through adaptive learning rate modulation.
\end{itemize}

Despite these promising results, CDASR faces challenges with extremely complex scenes involving reflections and semantic anomalies. Future work should focus on developing more advanced semantic parsing techniques and exploring the integration of temporal information for video super-resolution applications.

\section{Related Works}

SISR has been a longstanding challenge in computer vision, aiming to reconstruct high-resolution (HR) images from their low-resolution (LR) counterparts. This section provides an overview of recent advancements in deep learning-based SR methods, focusing on CNN and transformer-based approaches.

\subsection{CNN-based Super-Resolution}

Deep learning has revolutionized the field of SISR. Early CNN-based methods like SRCNN \cite{dong2015image} and VDSR \cite{kim2016accurate} demonstrated the potential of deep learning for SISR.
Subsequent research explored increasingly sophisticated architectures to enhance performance. Lim et al. proposed EDSR \cite{lim2017enhanced}, which significantly improved results by optimizing residual network structures for SR. The introduction of attention mechanisms marked another significant advancement. Zhang et al. developed RCAN \cite{zhang2018image}, incorporating channel attention to adaptively re-calibrate feature responses. This concept was further extended in HAN \cite{niu2020single}, which explored multi-scale attention across channels, layers, and spatial locations. Lu et al. proposed UnetSR \cite{lu2022single}, adapting the U-Net architecture \cite{ronneberger2015u} for SR tasks with a mixed gradient loss. 

Recognizing the limitations of local convolutions, researchers began investigating methods to capture long-range dependencies. Zhou et al. introduced IGNN \cite{zhou2020cross}, leveraging graph neural networks to model cross-scale patch recurrence. Dai et al. proposed SAN \cite{dai2019second}, employing second-order attention to enhance feature correlations. These approaches demonstrated the importance of non-local information in SR tasks. Mei et al. presented NLSA \cite{mei2021image}, combining non-local operations with sparse representation to balance global modeling capability and computational efficiency.

Recent works focused on developing lightweight SR models to address practical deployment constraints. Li et al. proposed LAPAR \cite{li2020lapar}, employing a linearly-assembled pixel-adaptive regression approach. Luo et al. introduced LatticeNet \cite{luo2020latticenet}, utilizing an efficient lattice block structure. Lu et al. developed DenseSR \cite{lu2022dense}, introducing a dense U-Net with shuffle pooling layer. These efforts highlight the ongoing challenge of balancing model complexity and performance in real-world SR applications.

\subsection{Transformer-based Super-Resolution}

The success of transformer architectures in natural language processing has inspired their adoption in computer vision tasks, including SR. Liang et al. introduced SwinIR \cite{liang2021swinir}, adapting the Swin Transformer for image restoration. This shift towards transformer-based models represents a promising direction for overcoming the inherent limitations of CNNs in capturing long-range dependencies. Building on this foundation, Chen et al. proposed IPT \cite{chen2021pre}, a pre-trained transformer model that demonstrates excellent performance across various low-level vision tasks. Lu et al. introduced EDT \cite{li2023efficient}, exploring efficient transformer-based image pre-training specifically tailored for low-level vision tasks. SwinFIR \cite{zhang2022swinfir} further refined the transformer-based approach by incorporating frequency domain information, while HAT \cite{chen2023activating} introduced a hybrid attention transformer to balance local and global feature interactions.

Recent advancements have focused on optimizing transformer architectures for SR tasks. ESRT \cite{lu2022transformer} introduced an efficient transformer structure that leverages multi-scale feature fusion. Hsu et al. proposed DRCT \cite{hsu2024drct}, a dense-residual-connected transformer that mitigates information bottlenecks in SR tasks through enhanced feature preservation and information flow stabilization. These approaches demonstrate the potential of transformer-based models to achieve state-of-the-art performance while addressing computational efficiency concerns. 

Modern advances in semantic-based approaches have shown promise for domain generalization. Liu et al. \cite{liu2024learning} proposed a domain-agnostic expert framework for person re-identification that transfers knowledge across unlabeled domains. Similarly, Chen et al. \cite{chen2024conjugated} introduced a conjugated semantic pool that improves out-of-distribution detection with pre-trained vision-language models. These works validate the effectiveness of semantic guidance for cross-domain generalization, aligning with our CLIP-based approach.

\section{Proposed Method}

To address the challenges of domain-adaptive super-resolution, we propose a novel framework that leverages CLIP's semantic understanding capabilities. Our method consists of three key components: a CLIP-guided feature alignment module, a domain-adaptive reconstruction network, and a meta-learning inspired few-shot adaptation strategy. This section details each component and their integration within the overall framework, providing insights into how they collectively enhance cross-domain generalization in single image super-resolution tasks.

\subsection{Overview}

Let $\mathcal{D}_s = \{(x_i^s, y_i^s)\}_{i=1}^{N_s}$ denote the source domain dataset used to pre-train our super-resolution model, where $x_i^s$ and $y_i^s$ represent low-resolution (LR) and high-resolution (HR) image pairs, respectively. Our goal is to adapt this model to $K$ target domains $\{\mathcal{D}_t^k\}_{k=1}^K$, each with a distinct scene, using only a few labeled examples per domain. We denote the few-shot target domain datasets as $\mathcal{D}_t^k = \{(x_i^{t,k}, y_i^{t,k})\}_{i=1}^{N_t^k}$, where $N_t^k \ll N_s$.

\begin{figure}[t]
	\centering
	\includegraphics[width=\linewidth]{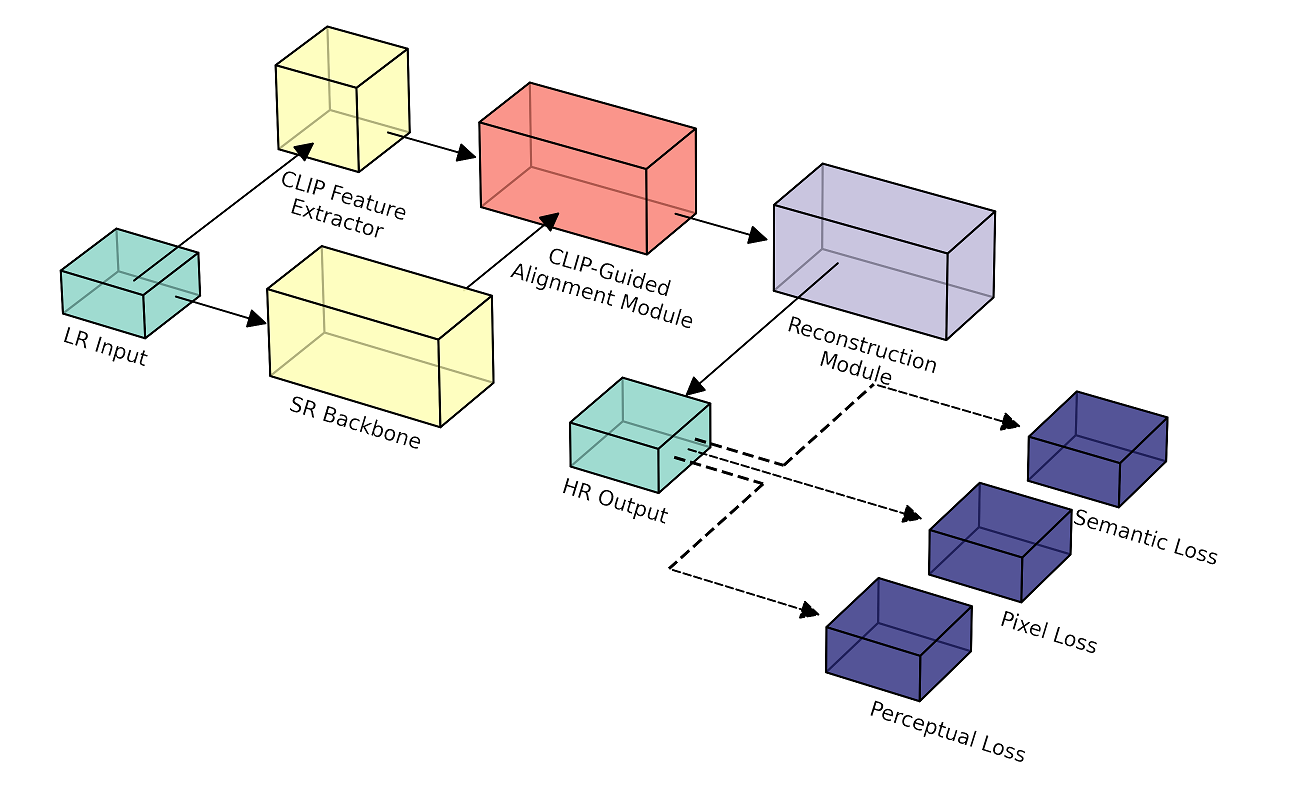}
	\caption{Overview of the proposed CDASR framework. The CLIP-guided alignment module (a) fuses semantic information from CLIP with SR features, while the reconstruction module (b) generates the final high-resolution output.}
	\label{fig:cdasr_framework}
\end{figure}

In Figure \ref{fig:cdasr_framework}, the CDASR framework operates in multiple stages. At first, the low-resolution input is simultaneously processed by the CLIP feature extractor ($E_{\text{CLIP}}$) and the super-resolution backbone ($F_{\theta}$). The extracted CLIP features and super-resolution features are then fused in the CLIP-Guided Alignment Module ($G_{\phi}$), which adapts the features to the target domain. The aligned features are subsequently fed into the reconstruction module $R_{\psi}$ to generate the high-resolution output. The entire process is optimized using a combination of perceptual, pixel-wise, and semantic loss functions, ensuring both low-level fidelity and high-level semantic consistency in the super-resolved images across diverse domains.

\subsection{CLIP-Guided Feature Alignment}

To efficiently introduce semantic information into super-resolution, we propose a novel CLIP-guided feature alignment mechanism. This approach leverages the rich semantic representations learned by pre-trained CLIP models to facilitate effective cross-domain knowledge transfer.

\begin{figure}[t]
	\centering
	\includegraphics[width=\linewidth]{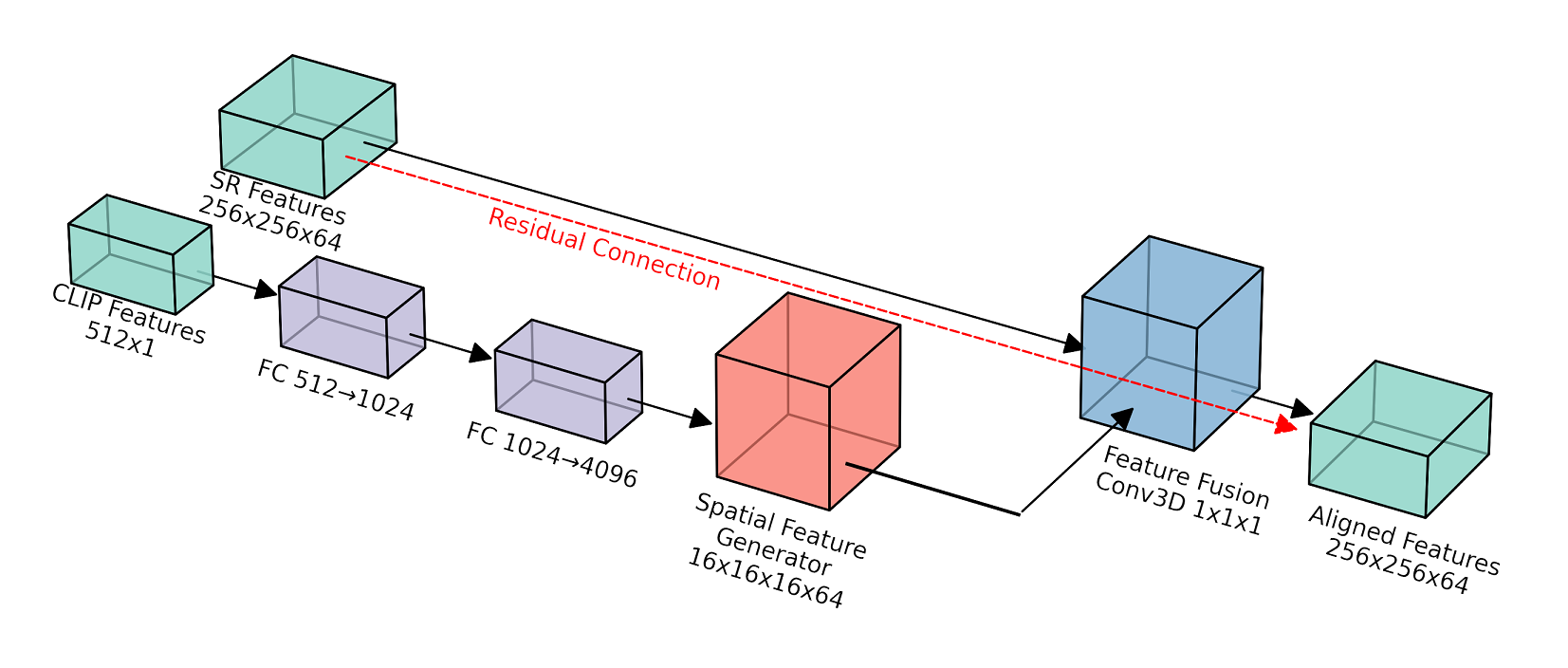}
	\caption{CLIP-guided alignment module fusing semantic and spatial information for enhanced super-resolution.}
	\label{fig:clip_module}
\end{figure}

The CLIP-guided alignment module, illustrated in Figure \ref{fig:clip_module}, integrates CLIP features with SR features through a sophisticated multi-stage process. Initially, CLIP features undergo a fully connected operation to reshape them into a spatial format. These spatially-aligned features are then processed through 3D convolutions. Simultaneously, the original SR features flow through a parallel path. The two streams converge in a fusion step, where the processed CLIP features are combined with the SR features. To further enhance information preservation, a residual connection reintroduces the original SR features. This intricate architecture culminates in the production of aligned features that effectively incorporate both semantic and spatial information.

This design leverages CLIP's semantic understanding while preserving spatial information crucial for super-resolution. The multi-stage approach allows for gradual integration of semantic and spatial features, potentially mitigating information loss. The residual connection ensures that fine-grained details from the original SR features are not overlooked, striking a balance between semantic enrichment and detail preservation.

\subsubsection{CLIP Feature Extraction}

We apply a fixed pre-trained CLIP model to extract domain-agnostic semantic features from both source and target domain images. The CLIP model, developed by Radford et al. \cite{radford2021learning}, has demonstrated remarkable zero-shot transfer capabilities across various vision tasks. In the proposed framework, we specifically employ the vision encoder of CLIP, which remains frozen throughout our training process.

Given an image $I$, we define the CLIP feature extraction process as:
\begin{equation}
	f_{\text{CLIP}}(I) = E_{\text{CLIP}}(I) / |E_{\text{CLIP}}(I)|_2
\end{equation}
where $E_{\text{CLIP}}$ is the fixed CLIP image encoder, and $|\cdot|_2$ denotes L2 normalization. These normalized CLIP features capture high-level semantic information that is largely invariant to low-level domain-specific characteristics.

\subsubsection{Domain-Adaptive Feature Alignment Module}

We introduce a domain-adaptive alignment module $G_{\phi}$ to align feature distributions across domains. This module processes CLIP features $f_{\text{CLIP}}$ and super-resolution features $F_{\theta}(x)$ through multi-stage transformations:
\begin{equation}
	G_{\phi}: [F_{\theta}(x), f_{\text{CLIP}}(x)] \rightarrow \mathbb{R}^d
\end{equation}
where $d$ is the aligned feature space dimension. The module consists of three key components: CLIP feature processing, spatial feature generation, and feature fusion.

The CLIP feature processor transforms the input CLIP features through a two-layer MLP with ReLU activation:

\begin{equation}
	f_{\text{proc}} = \text{ReLU}(W_2 \cdot \text{ReLU}(W_1 \cdot f_{\text{CLIP}} + b_1) + b_2)
\end{equation}
where $W_1 \in \mathbb{R}^{h \times c}$, $W_2 \in \mathbb{R}^{s \times h}$, with hidden dimension $h=1024$ and output dimension $s=512$. The processed features are then expanded spatially and passed through convolutional layers to generate spatial features:
\begin{equation}
	f_{\text{spatial}} = \text{Conv}_2(\text{ReLU}(\text{Conv}_1(f_{\text{proc}})))
\end{equation}

Finally, these spatial features are fused with the original super-resolution features:
\begin{equation}
	f_{\text{aligned}} = \text{Conv}{\text{fusion}}([F_{\theta}(x), f_{\text{spatial}}])
\end{equation}

This architecture enables effective domain adaptation by leveraging CLIP's semantic understanding while preserving spatial information crucial for super-resolution tasks. The use of residual connections and layer normalization further enhances feature propagation and stability during training.

\subsubsection{Training Objective}

We employ a multi-component loss function to train the novel alignment module and the super-resolution model:
\begin{equation}
	\mathcal{L}_{\text{total}} = \lambda_{\text{pixel}} \mathcal{L}_{\text{pixel}} + \lambda_{\text{perc}} \mathcal{L}_{\text{perc}} + \lambda_{\text{sem}} \mathcal{L}_{\text{sem}}
\end{equation}
where $\mathcal{L}_{\text{pixel}}$ is the pixel-wise L1 loss, $\mathcal{L}_{\text{perc}}$ is the perceptual loss\cite{johnson2016perceptual}, and $\mathcal{L}_{\text{sem}}$ is the semantic consistency loss. The semantic consistency loss is defined as:
\begin{equation}
	\mathcal{L}_{\text{sem}} = |f_{\text{CLIP}}(F_{\theta}(x_{LR})) - f_{\text{CLIP}}(x_{HR})|_2^2
\end{equation}

This loss function encourages the model to generate high-resolution images that are not only close to the target at the pixel level but also consistent in terms of higher-level semantic features.

\subsubsection{Reconstruction Module}

\begin{figure}[t]
	\centering
	\includegraphics[width=\linewidth]{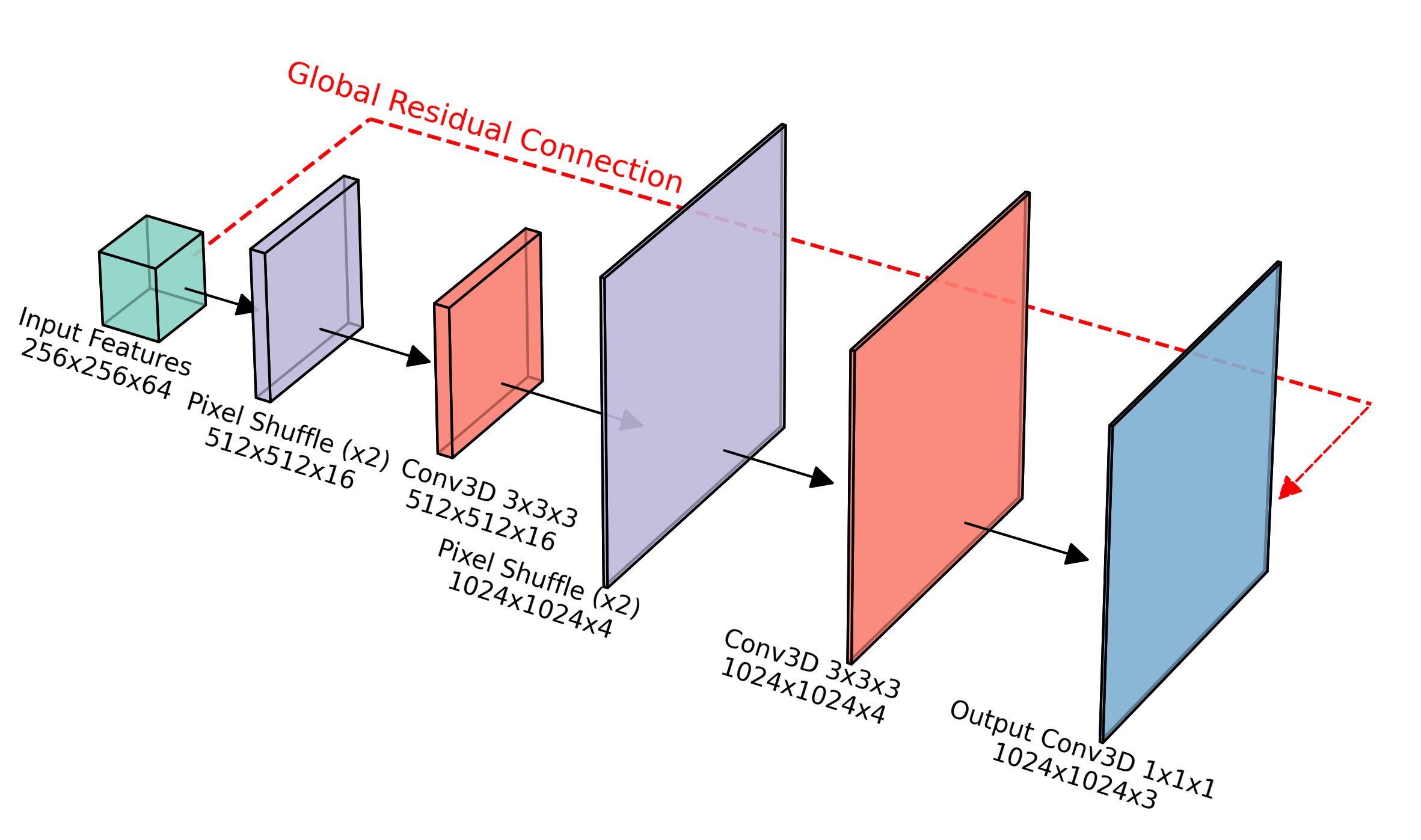}
	\caption{Reconstruction module architecture for generating high-resolution output from aligned features.}
	\label{fig:recons_mudule}
\end{figure}

The reconstruction module serves as the final stage in CDASR framework, responsible for generating the high-resolution output. As shown in Figure. \ref{fig:recons_mudule}, this module consists of a multi-stage upsampling layers followed by convolutional operations. The module takes the aligned features from the feature alignment module as input and progressively increases the spatial resolution while refining the details. The residual connection facilitate the learning of high-frequency information.

The architecture of the reconstruction module is designed to be scale-aware, adapting its structure based on the target upscaling factor. For $\times2$ and $\times4$ upscaling, we use a cascade of pixel shuffle layers, each followed by a 3$\times$3 convolutional layer. For $\times8$ and higher factors, we introduce an additional set of residual blocks between upsampling stages to enhance the model's capacity for detail synthesis. The final output is generated through a global residual connection, combining the upscaled features with the initial bicubic interpolation of the input, ensuring stability in training and preservation of low-frequency content.

\subsection{Optimization Procedure}

\subsubsection{Training Objective}

We employ a multi-component loss function to train the enhanced alignment module and the super-resolution model:

\begin{equation}
	\mathcal{L}_{\text{total}} = \mathcal{L}_{\text{1}} + \lambda_{\text{perc}} \mathcal{L}_{\text{perc}} + \lambda_{\text{sem}} \mathcal{L}_{\text{sem}}
\end{equation}
where $\mathcal{L}_{\text{1}}$ is the standard pixel-wise L1 loss, $\mathcal{L}_{\text{perc}}$ is the perceptual loss, and $\mathcal{L}_{\text{sem}}$ is the novel Multi-Scale Semantic Consistency Loss.

The CLIP-based semantic consistency loss $\mathcal{L}_{\text{sem}}$ is defined as:

\begin{equation}
	\mathcal{L}_{\text{sem}} = \|f_{\text{CLIP}}(F_{\theta}(x_{LR})) - f_{\text{CLIP}}(x_{HR})\|_2^2
\end{equation}
where $x_{LR}$ and $x_{HR}$ are the low-resolution and high-resolution images respectively, $F_{\theta}$ is the super-resolution model, and $f_{\text{CLIP}}$ is the CLIP feature extractor.

The purpose of this loss function is to minimize the distance between the super-resolution result and the true high-resolution image in the CLIP feature space, thereby ensuring semantic consistency throughout the super-resolution process. By doing so, we encourage the model to generate high-resolution images that are not only close to the target at the pixel level but also consistent in terms of higher-level semantic features.

The training process of the proposed CDASR framework consists of three main stages: pre-training, CLIP-guided alignment, and meta-learning. The inference process involves adapting the model to heterogeneous target domains using few-shot samples. Algorithm \ref{alg:train_inference} provides a detailed description of our training and inference procedures.

\subsubsection{Training and Inference}

The training and inference procedures for the proposed CDASR framework are detailed in Algorithm \ref{alg:train_inference}. We show the algorithm to reflect the implementation  steps:

\begin{algorithm}
	\caption{CDASR Training Procedure}
	\label{alg:train_inference}
	\begin{algorithmic}[1]
		\REQUIRE Source domain dataset $\mathcal{D}_s$, CLIP model $E_{\text{CLIP}}$
		\ENSURE Optimized super-resolution model for source domains
		
		\STATE \textbf{function} TRAIN($\mathcal{D}_s, E_{\text{CLIP}}$)
		\STATE Initialize super-resolution backbone $F_{\theta}$
		\STATE Initialize domain-adaptive alignment module $G_{\phi}$
		\STATE Initialize reconstruction module $R_{\psi}$
		\STATE Initialize meta-learner $M$
		\STATE Initialize optimizer with learning rate $\eta$
		\STATE Initialize learning rate scheduler
		
		\FOR{$epoch = 1$ to $N_{epochs}$}
		\FOR{each batch $(x_{LR}, x_{HR})$ in $\mathcal{D}_s$}
		\STATE Extract CLIP features: $f_{\text{CLIP}} = E_{\text{CLIP}}(x_{LR})$
		\STATE Compute SR features: $f_{SR} = F_{\theta}(x_{LR})$
		\STATE Compute aligned features: \\$f_{\text{aligned}} = G_{\phi}([f_{SR}, f_{\text{CLIP}}])$
		\STATE Generate SR images: $x_{SR} = R_{\psi}(f_{\text{aligned}})$
		\STATE Compute $\mathcal{L}_{\text{total}} = \mathcal{L}_{\text{1}} + \lambda_{\text{perc}} \mathcal{L}_{\text{perc}} + \lambda_{\text{sem}} \mathcal{L}_{\text{sem}}$
		\STATE Update $F_{\theta}$, $G_{\phi}$, and $R_{\psi}$ by minimizing $\mathcal{L}_{\text{total}}$
		\ENDFOR
		\STATE Update learning rate using scheduler
		\ENDFOR
	\end{algorithmic}
\end{algorithm}

The training procedure of the proposed CDASR framework consists of three main stages: pre-training, CLIP-guided alignment, and meta-learning. Algorithm \ref{alg:train_inference} provides a detailed description of our training and inference procedures. During training, we initialize the super-resolution backbone $F_{\theta}$, the domain-adaptive alignment module $G_{\phi}$, the reconstruction module $R_{\psi}$, and the meta-learner $M$. For each batch, we extract CLIP features, compute super-resolution features, align these features using $G_{\phi}$, and generate super-resolved images using the reconstruction module $R_{\psi}$. The model is updated by minimizing a multi-component loss function that includes L1 loss, perceptual loss, and semantic consistency loss.

\subsubsection{Meta-Learning-based Few-Shot Adaptation}

We propose a meta-learning inspired few-shot adaptation strategy that complements the CLIP-guided alignment for efficient adaptation to new target domains\cite{finn2017model}. This approach is applied during the ADAPT function in Algorithm 1, allowing the model to leverage both general semantic knowledge and domain-specific information efficiently.

To enhance the efficiency of few-shot adaptation, we introduce an adaptive learning rate modulation scheme within the ADAPT function, following principles of meta-transfer learning \cite{sun2019meta}. For each parameter $\theta_i$ in the model (including $F_{\theta}$, $G_{\phi}$, and $R_{\psi}$), we learn a corresponding meta-learning rate $\alpha_i$ that determines the adaptability:

\begin{equation}
	\theta_i' = \theta_i - \alpha_i \nabla_{\theta_i} \mathcal{L}_{\mathcal{D}_t^k}(\theta)
\end{equation}
where $\mathcal{L}_{\mathcal{D}_t^k}$ is the loss computed on the support set $\mathcal{S}$ from the new target domain $\mathcal{D}_t^k$.

The meta-learning rates $\alpha_i$ are updated using the meta-objective:
\begin{equation}
	\alpha_i \leftarrow \alpha_i - \gamma \frac{\partial}{\partial \alpha_i} \mathcal{L}_{\mathcal{Q}}(\theta')
\end{equation}
where $\gamma$ is a higher-order learning rate, and $\mathcal{L}_{\mathcal{Q}}$ is the loss computed on the query set $\mathcal{Q}$ from the target domain.

This adaptive scheme allows the model to learn which parameters should be more plastic during few-shot adaptation, leading to effective domain transfer without affecting the original training process.

\begin{algorithm}
	\caption{CDASR Adaptation Procedure}
	\label{alg:Adaptation}
	\begin{algorithmic}[1]	
		\STATE \textbf{function} ADAPT($F_{\theta}, G_{\phi}, R_{\psi}, \mathcal{D}_t^{new}, E_{\text{CLIP}}$)
		\STATE Initialize meta-learner $M$ with current model parameters
		\FOR{$episode = 1$ to $N_{episodes}$}
		\STATE Sample support set $\mathcal{S}$ and query set $\mathcal{Q}$ from $\mathcal{D}_t^{new}$
		\STATE $\theta', \phi', \psi' = M.adapt(\mathcal{S}, F_{\theta}, G_{\phi}, R_{\psi})$
		\FOR{each batch $(x_{LR}, x_{HR})$ in $\mathcal{Q}$}
		\STATE Extract CLIP features: $f_{\text{CLIP}} = E_{\text{CLIP}}(x_{LR})$
		\STATE Compute SR features: $f_{SR} = F_{\theta'}(x_{LR})$
		\STATE Compute aligned features: \\$f_{\text{aligned}} = G_{\phi'}([f_{SR}, f_{\text{CLIP}}])$
		\STATE Generate SR images: $x_{SR} = R_{\psi'}(f_{\text{aligned}})$
		\STATE Compute loss $\mathcal{L}$ on $(x_{SR}, x_{HR})$
		\STATE Update $M$ using $\mathcal{L}$ and MAML or Reptile algorithm
		\ENDFOR
		\ENDFOR
		\RETURN Adapted $F_{\theta'}, G_{\phi'}, R_{\psi'}$
	\end{algorithmic}
\end{algorithm}

Algorithm \ref{alg:Adaptation} outlines the adaptation procedure for CDASR. It initializes a meta-learner with current model parameters and iterates through episodes, sampling support and query sets from the target domain. The meta-learner adapts model components using the support set. For each query batch, CLIP features and SR features are extracted, aligned, and used to generate SR images. The loss is computed and applied to update the meta-learner. This process enables efficient adaptation to target domains with limited samples, leveraging the pre-trained model's knowledge.

\subsection{Theoretical Analysis}

We present a theoretical analysis of the proposed CDASR method, providing insights into its generalization capabilities across diverse domains and scaling factors\cite{ben2010theory}.

We begin with the standard domain adaptation bound:

\begin{equation}
	R_t(h) \leq R_s(h) + \frac{1}{2}d_{\mathcal{H}\Delta\mathcal{H}}(P_s, P_t) + \lambda
\end{equation}
where $R_t(h)$ and $R_s(h)$ are the true risks on the target and source domains respectively, $d_{\mathcal{H}\Delta\mathcal{H}}(P_s, P_t)$ is the $\mathcal{H}\Delta\mathcal{H}$-divergence between source and target distributions, and $\lambda$ is the combined error of the ideal joint hypothesis.

Next, we bound $R_s(h)$ using the Rademacher complexity\cite{bartlett2002rademacher}:

\begin{equation}
	R_s(h) \leq \hat{R}_s(h) + 2\mathfrak{R}_s(\mathcal{H}_{\text{CLIP}}) + 3\sqrt{\frac{\log(2/\delta)}{2m}}
\end{equation}
This bound holds with probability at least $1-\delta$ over the choice of $m$ samples. Here, $\hat{R}_s(h)$ is the empirical risk on the source domain, and $\mathfrak{R}_s(\mathcal{H}_{\text{CLIP}})$ is the Rademacher complexity of $\mathcal{H}_{\text{CLIP}}$ on the source domain.

To incorporate the CLIP-guided alignment, we upper-bound the $\mathcal{H}\Delta\mathcal{H}$-divergence using the Maximum Mean Discrepancy (MMD)\cite{long2015learning} in the CLIP space:

\begin{equation}
	d_{\mathcal{H}\Delta\mathcal{H}}(P_s, P_t) \leq 2 \cdot \text{MMD}_{\text{CLIP}}(P_s, P_t)
\end{equation}

For the meta-learning component, we introduce an expected divergence term to replace $\lambda$:

\begin{equation}
	\lambda \leq \lambda_2 \cdot \mathbb{E}_{\tau}[d_{\mathcal{H}_{\text{CLIP}}}(P_{\tau}, P_t)]
\end{equation}
where $\mathbb{E}_{\tau}[d_{\mathcal{H}_{\text{CLIP}}}(P_{\tau}, P_t)]$ is the expected divergence between task-specific and target distributions.

To account for the few-shot adaptation, we introduce a more comprehensive analysis. Let $S_t = \{(x_i, y_i)\}_{i=1}^n$ be a small sample of $n$ labeled examples from the target domain. We define the few-shot adaptation loss as:

\begin{equation}
	\mathcal{L}_{\text{few-shot}}(h, S_t) = \frac{1}{n} \sum_{i=1}^n \ell(h(x_i), y_i)
\end{equation}
where $\ell$ is a suitable loss function. To bound the generalization error of the few-shot adapted model, we employ the meta-learning theory developed by Finn et al. \cite{finn2019online}. Let $h_{\theta}$ be the initial model parameters and $h_{\theta'}$ be the adapted model after $k$ gradient steps on $S_t$. We have:

\begin{equation}
	\begin{split}
		R_t(h_{\theta'}) \leq & R_t(h_{\theta}) + \mathcal{L}_{\text{few-shot}}(h_{\theta'}, S_t) \\
		& + 2\mathfrak{R}_n(\mathcal{H}_{\text{CLIP}}) + 3\sqrt{\frac{\log(2/\delta)}{2n}}
	\end{split}
\end{equation}
where $\mathfrak{R}_n(\mathcal{H}_{\text{CLIP}})$ is the Rademacher complexity\cite{bartlett2002rademacher} of $\mathcal{H}_{\text{CLIP}}$ on the $n$ samples from the target domain.

Combining all these bounds and introducing trade-off parameters $\lambda_1$, $\lambda_2$, and $\lambda_3$, we arrive at the final bound stated in the following theorem.

\textbf{Theorem 1 (CLIP-Guided Generalization Bound):}
Let $\mathcal{H}_{\text{CLIP}}$ be a hypothesis class operating on CLIP embeddings. For any $h_{\theta} \in \mathcal{H}_{\text{CLIP}}$, its few-shot adapted version $h_{\theta'}$, and $\delta > 0$, with probability at least $1-\delta$ over the choice of samples, we have:

\begin{equation}
	\begin{split}
		R_t(h_{\theta'}) \leq & \hat{R}_s(h_{\theta}) + 2\mathfrak{R}_s(\mathcal{H}_{\text{CLIP}}) + 3\sqrt{\frac{\log(2/\delta)}{2m}} \\
		& + \lambda_1 \cdot \text{MMD}_{\text{CLIP}}(P_s, P_t) \\
		& + \lambda_2 \cdot \mathbb{E}_{\tau}[d_{\mathcal{H}_{\text{CLIP}}}(P_{\tau}, P_t)] \\
		& + \lambda_3 \cdot (\mathcal{L}_{\text{few-shot}}(h_{\theta'}, S_t) + 2\mathfrak{R}_n(\mathcal{H}_{\text{CLIP}}) + 3\sqrt{\frac{\log(2/\delta)}{2n}})
	\end{split}
\end{equation}
where $\lambda_1, \lambda_2, \lambda_3 > 0$ are trade-off parameters, $m$ is the number of source domain samples, and $n$ is the number of target domain samples used in few-shot adaptation.

This bound provides a theoretical justification for the CDASR method, showing how it leverages CLIP embeddings, meta-learning, and few-shot adaptation to improve generalization across diverse domains and scaling factors. The Rademacher complexity terms $\mathfrak{R}_s(\mathcal{H}_{\text{CLIP}})$ and $\mathfrak{R}_n(\mathcal{H}_{\text{CLIP}})$ capture the richness of the hypothesis class in the CLIP embedding space for both source and target domains. The MMD term $\text{MMD}_{\text{CLIP}}(P_s, P_t)$ quantifies the alignment between source and target domains. The expected divergence $\mathbb{E}_{\tau}[d_{\mathcal{H}_{\text{CLIP}}}(P_{\tau}, P_t)]$ accounts for the meta-learning aspect, while the few-shot loss $\mathcal{L}_{\text{few-shot}}(h_{\theta'}, S_t)$ and its associated complexity terms represent the model's ability to adapt quickly with limited target domain samples.

\section{Experimental Results}

We conduct extensive experiments to evaluate the proposed CLIP-aware domain-sdaptive super-resolution method. The proposed experiments focus on three key aspects: (1) comparison with state-of-the-art methods, (2) ablation studies to validate novel modules, and (3) analysis of the domain adaptation capabilities.

\subsection{Experimental setting}
\subsubsection{Datasets}

We train our CDASR model on DF2K, a combination of DIV2K \cite{timofte2017ntire} (800 training images) and Flickr2K \cite{wang2019flickr1024} (2650 images). For evaluation, we use standard SISR benchmark datasets: Set5 \cite{bevilacqua2012low}, Set14 \cite{zeyde2010single}, BSD100 \cite{martin2001database}, Urban100 \cite{huang2015single}, and Manga109 \cite{matsui2017sketch}. Low-resolution images are generated using bicubic downsampling with scaling factors of 2, 4, and 8.

\subsubsection{Implementation Details}

We implement our framework using PyTorch 1.9.0 \cite{paszke2019pytorch}. The super-resolution backbone $F_{\theta}$ is based on the EDSR architecture \cite{lim2017enhanced} with modifications to incorporate the aligned features from $G_{\phi}$. We use the ViT-B/32 variant of CLIP \cite{radford2021learning} for feature extraction.

During training, we use Adam optimizer with a learning rate of $1\times10^{-4}$. The batch size is set to 16 for pre-training and 4 for meta-learning, with 5-shot learning for each target domain. We set $\lambda_{\text{align}}=0.1$ and $\lambda_{\text{scale}}=0.01$ based on validation performance. The project code is publicly available online.

\subsection{Comparison to state-of-the-arts methods}

We compare the proposed CDASR method with state-of-the-art SISR methods, including CNN-based approaches (EDSR \cite{lim2017enhanced}, RCAN \cite{zhang2018image}, SAN \cite{dai2019second}, IGNN \cite{zhou2020cross}, HAN \cite{niu2020single}, NLSA\cite{mei2021image}) and Transformer-based methods (SwinIR \cite{liang2021swinir}, EDT \cite{li2023efficient}). 

Table \ref{tab:comparison_results} presents quantitative results on benchmark datasets for $\times2$ and $\times4$ upscaling factors. The DIV2K dataset was used to train the CNN-based methods EDSR, RCAN, SAN, IGNN, RNAN, HAN, and NLSA. Networks based on Vision Transformer include SwinIR and EDT. EDT is trained on the ImageNet dataset and EDT$^\dagger$ is fine-tuned on the DF2K dataset to get optimal performance. And SwinIR only is trained on the DF2K dataset. Following EDT, our SwinFIR is first trained on the ImageNet \cite{deng2009imagenet}, and then fine-tuned on the DF2K dataset.

In Table \ref{tab:comparison_results}, the proposed CDASR demonstrates consistent superior performance across various datasets and scaling factors. For $\times2$ upscaling, CDASR achieves notable improvements, particularly on challenging datasets like Urban100 and Manga109, with PSNR gains of 0.01dB and 0.05dB respectively compared to the next best method (excluding EDT$^\dagger$). At $\times4$ scaling, CDASR's advantages become more pronounced, outperforming other methods by margins of 0.06dB on Urban100 and 0.02dB on Manga109. The most significant improvements are observed at $\times8$ scaling, where CDASR surpasses the previous best results by 0.06dB on BSD100, 0.07dB on Urban100, and 0.11dB on Manga109. Notably, CDASR consistently outperforms SwinIR and EDT across all scales and datasets, with the performance gap widening as the scaling factor increases. These results underscore CDASR's effectiveness with diverse image types and the particular strength in challenging scenarios.

\begin{table*}[!t]
	\centering
	\caption{Quantitative comparison with state-of-the-art methods on benchmark datasets for classical image super-resolution. The best and second-best results are marked in \textcolor{red}{red} and \textcolor{blue}{blue}, respectively.}
	\label{tab:comparison_results}
	\resizebox{\textwidth}{!}{%
		\begin{tabular}{@{}l c c c c c c c c c c c c@{}}
			\toprule
			\multirow{2}{*}{Method} & \multirow{2}{*}{Scale} & \multirow{2}{*}{Training Dataset} &
			\multicolumn{2}{c}{Set5} & \multicolumn{2}{c}{Set14} & \multicolumn{2}{c}{BSD100} & \multicolumn{2}{c}{Urban100} & \multicolumn{2}{c}{Manga109} \\
			\cmidrule(lr){4-5} \cmidrule(lr){6-7} \cmidrule(lr){8-9} \cmidrule(lr){10-11} \cmidrule(lr){12-13}
			& & & PSNR & SSIM & PSNR & SSIM & PSNR & SSIM & PSNR & SSIM & PSNR & SSIM \\
			\midrule
			EDSR \cite{lim2017enhanced} & $\times2$ & DIV2K & 38.11 & 0.9602 & 33.92 & 0.9195 & 32.32 & 0.9013 & 32.93 & 0.9351 & 39.10 & 0.9773 \\
			RCAN \cite{zhang2018image} & $\times2$ & DIV2K & 38.27 & 0.9614 & 34.12 & 0.9216 & 32.41 & 0.9027 & 33.34 & 0.9384 & 39.44 & 0.9786 \\
			SAN \cite{dai2019second} & $\times2$ & DIV2K & 38.31 & 0.9620 & 34.07 & 0.9213 & 32.42 & 0.9028 & 33.10 & 0.9370 & 39.32 & 0.9792 \\
			IGNN \cite{zhou2020cross} & $\times2$ & DIV2K & 38.24 & 0.9613 & 34.07 & 0.9217 & 32.41 & 0.9025 & 33.23 & 0.9383 & 39.35 & 0.9786 \\
			HAN \cite{niu2020single} & $\times2$ & DIV2K & 38.27 & 0.9614 & 34.16 & 0.9217 & 32.41 & 0.9027 & 33.35 & 0.9385 & 39.46 & 0.9785 \\
			NLSN \cite{mei2021image} & $\times2$ & DIV2K & 38.34 & 0.9618 & 34.08 & 0.9231 & 32.43 & 0.9027 & 33.42 & 0.9394 & 39.59 & 0.9789 \\
			SwinIR  \cite{mei2021image} & $\times2$ & DF2K & \textcolor{blue}{38.42} & \textcolor{blue}{0.9623} & 34.46 & 0.9250 & 32.53 & 0.9041 & 33.81 & 0.9427 & 39.92 & 0.9797 \\
			EDT \cite{li2023efficient} & $\times2$ & DF2K & 38.39 & 0.9610 & \textcolor{blue}{34.57} & \textcolor{blue}{0.9258} & \textcolor{blue}{32.52} & \textcolor{blue}{0.9041} & 33.80 & 0.9425 & 39.93 & 0.9800 \\
			EDT$^\dagger$ \cite{li2023efficient} & $\times2$ & DF2K & \textcolor{red}{38.63} & \textcolor{red}{0.9632} & \textcolor{red}{34.80} & \textcolor{red}{0.9273} & \textcolor{red}{32.62} & \textcolor{red}{0.9052} & \textcolor{red}{34.27} & \textcolor{red}{0.9456} & \textcolor{red}{40.37} & \textcolor{red}{0.9811} \\
			\textbf{CDASR} & $\times2$ & DF2K & 38.46 & 0.9619 & 34.50 & 0.9247 & 32.53 & 0.9038 & \textcolor{blue}{33.82} & \textcolor{blue}{0.9428} & \textcolor{blue}{39.97} & \textcolor{blue}{0.9808} \\
			\midrule
			EDSR \cite{lim2017enhanced} & $\times4$ & DIV2K & 32.46 & 0.8968 & 28.80 & 0.7876 & 27.71 & 0.7420 & 26.64 & 0.8033 & 31.02 & 0.9148 \\
			RCAN \cite{zhang2018image} & $\times4$ & DIV2K & 32.63 & 0.9002 & 28.87 & 0.7889 & 27.77 & 0.7436 & 26.82 & 0.8087 & 31.22 & 0.9173 \\
			SAN \cite{dai2019second} & $\times4$ & DIV2K & 32.64 & 0.9003 & 28.92 & 0.7888 & 27.78 & 0.7436 & 26.79 & 0.8068 & 31.18 & 0.9169 \\
			IGNN \cite{zhou2020cross} & $\times4$ & DIV2K & 32.57 & 0.8998 & 28.85 & 0.7891 & 27.77 & 0.7434 & 26.84 & 0.8090 & 31.28 & 0.9182 \\
			HAN \cite{niu2020single} & $\times4$ & DIV2K & 32.64 & 0.9002 & 28.90 & 0.7890 & 27.80 & 0.7442 & 26.85 & 0.8094 & 31.42 & 0.9177 \\
			NLSN \cite{mei2021image} & $\times4$ & DIV2K & 32.59 & 0.9000 & 28.87 & 0.7891 & 27.78 & 0.7444 & 26.96 & 0.8109 & 31.27 & 0.9184 \\
			SwinIR \cite{mei2021image} & $\times4$ & DF2K & 32.92 & 0.9044 & \textcolor{blue}{29.11} & \textcolor{blue}{0.7956} & 27.92 & 0.7489 & 27.45 & 0.8254 & 32.03 & 0.9260 \\
			EDT \cite{li2023efficient} & $\times4$ & DF2K & 32.82 & 0.9031 & 29.09 & 0.7939 & 27.91 & 0.7483 & 27.46 & 0.8246 & 32.05 & 0.9254 \\
			EDT$^\dagger$ \cite{li2023efficient} & $\times4$ & DF2K & \textcolor{red}{33.06} & \textcolor{red}{0.9055} & \textcolor{red}{29.23} & \textcolor{red}{0.7971} & \textcolor{red}{27.99} & \textcolor{red}{0.7510} & \textcolor{blue}{27.75} & \textcolor{blue}{0.8317} & \textcolor{blue}{32.39} & \textcolor{blue}{0.9283} \\
			\textbf{CDASR} & $\times4$ & DF2K & \textcolor{blue}{33.04} & \textcolor{blue}{0.9053} & 29.08 & 0.7950 & \textcolor{blue}{27.98} & \textcolor{blue}{0.7508} & \textcolor{red}{27.81} & \textcolor{red}{0.8321} & \textcolor{red}{32.41} & \textcolor{red}{0.9289} \\
			\midrule
			EDSR \cite{lim2017enhanced} & $\times8$ & DIV2K & 27.71 & 0.8042 & 24.94 & 0.6489 & 24.80 & 0.6192 & 22.47 & 0.6406 & 25.34 & 0.8149 \\
			RCAN \cite{zhang2018image} & $\times8$ & DIV2K & 27.84 & 0.8073 & 25.01 & 0.6512 & 24.86 & 0.6214 & 22.63 & 0.6485 & 25.61 & 0.8236 \\
			SAN \cite{dai2019second} & $\times8$ & DIV2K & 27.86 & 0.8077 & 25.03 & 0.6516 & 24.88 & 0.6218 & 22.65 & 0.6495 & 25.65 & 0.8248 \\
			IGNN \cite{zhou2020cross} & $\times8$ & DIV2K & 27.82 & 0.8068 & 25.00 & 0.6509 & 24.85 & 0.6211 & 22.61 & 0.6479 & 25.58 & 0.8228 \\
			HAN \cite{niu2020single} & $\times8$ & DIV2K & 27.85 & 0.8075 & 25.02 & 0.6514 & 24.87 & 0.6216 & 22.64 & 0.6490 & 25.63 & 0.8242 \\
			NLSN \cite{mei2021image} & $\times8$ & DIV2K & 27.88 & 0.8081 & 25.05 & 0.6520 & 24.89 & 0.6221 & 22.68 & 0.6505 & 25.69 & 0.8259 \\
			SwinIR \cite{mei2021image} & $\times8$ & DF2K & 28.05 & 0.8123 & 25.18 & 0.6561 & 24.98 & 0.6258 & 22.91 & 0.6600 & 26.05 & 0.8366 \\
			EDT \cite{li2023efficient} & $\times8$ & DF2K & 28.01 & 0.8115 & 25.16 & 0.6555 & 24.97 & 0.6254 & 22.89 & 0.6593 & 26.01 & 0.8357 \\
			EDT$^\dagger$ \cite{li2023efficient} & $\times8$ & DF2K & \textcolor{blue}{28.15} & \textcolor{blue}{0.8147} & \textcolor{blue}{25.26} & \textcolor{blue}{0.6585} & \textcolor{blue}{25.04} & \textcolor{blue}{0.6280} & \textcolor{blue}{23.05} & \textcolor{blue}{0.6652} & \textcolor{blue}{26.28} & \textcolor{blue}{0.8422} \\
			\textbf{CDASR} & $\times8$ & DF2K & \textcolor{red}{28.21} & \textcolor{red}{0.8159} & \textcolor{red}{25.31} & \textcolor{red}{0.6596} & \textcolor{red}{25.08} & \textcolor{red}{0.6291} & \textcolor{red}{23.12} & \textcolor{red}{0.6673} & \textcolor{red}{26.39} & \textcolor{red}{0.8445} \\
			\bottomrule
		\end{tabular}%
	}
\end{table*}

\begin{figure*}[t]
	\centering
	\includegraphics[width=\linewidth]{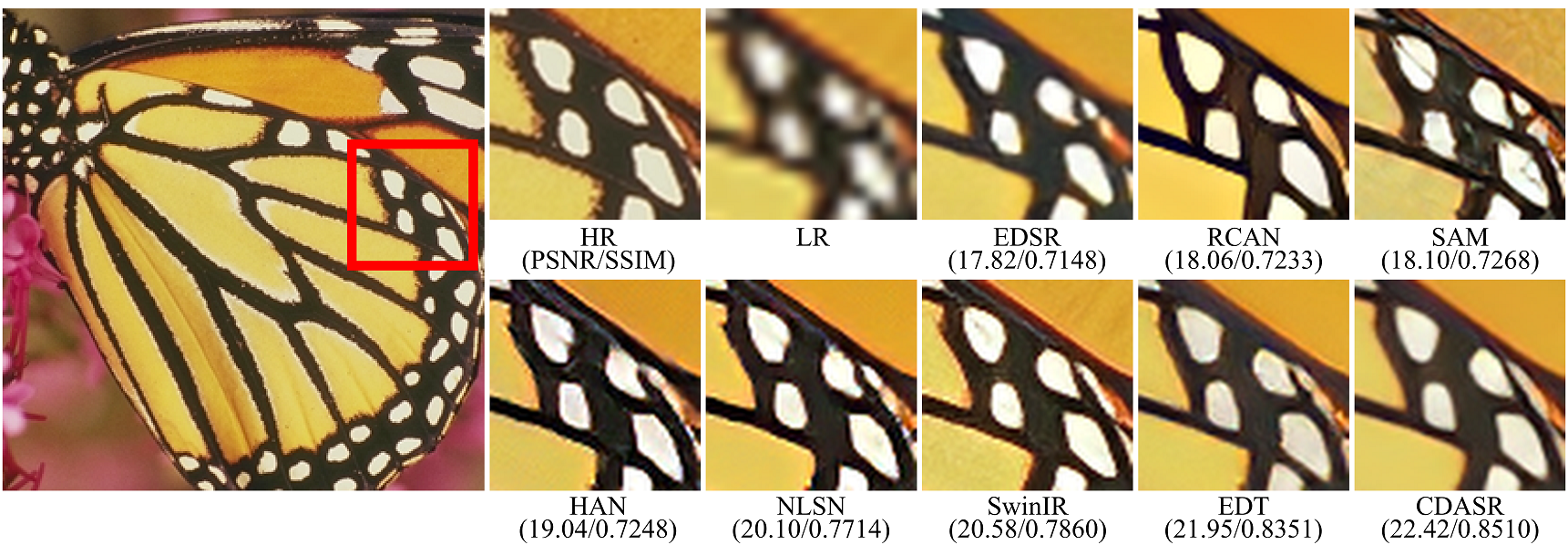}
	\caption{\textcolor{blue}{Visual comparison of super-resolution results ($\times4$) on the butterfly image from Set5 dataset. The red box highlights the detailed region shown across methods. PSNR/SSIM values beneath each result demonstrate CDASR's superior performance (22.42/0.8510) in preserving the intricate butterfly wing patterns.}}
	\label{fig:visual_comparison_1}
\end{figure*}

Figure \ref{fig:visual_comparison_1} showcases the super-resolution results ($\times4$) from the Set5 dataset, highlighting the effectiveness of CDASR in preserving intricate textures. While the reconstructed butterfly wing patterns by CDASR may not appear exceptionally sharp, they exhibit remarkable completeness in overall detail, closely aligning with the ground truth. Most existing methods struggle with accurate reconstruction at the junction points of fine black patterns, often introducing distortions. Although both RCAN and CDASR produce reasonable reconstructions with plausible details, CDASR's output demonstrates superior fidelity to the actual scene. This performance can be attributed to CDASR's novel CLIP-guided feature alignment mechanism, which enables the model to capture and preserve semantic information crucial for accurate texture reconstruction in complex natural scenes.

\begin{figure*}[t]
	\centering
	\includegraphics[width=\linewidth]{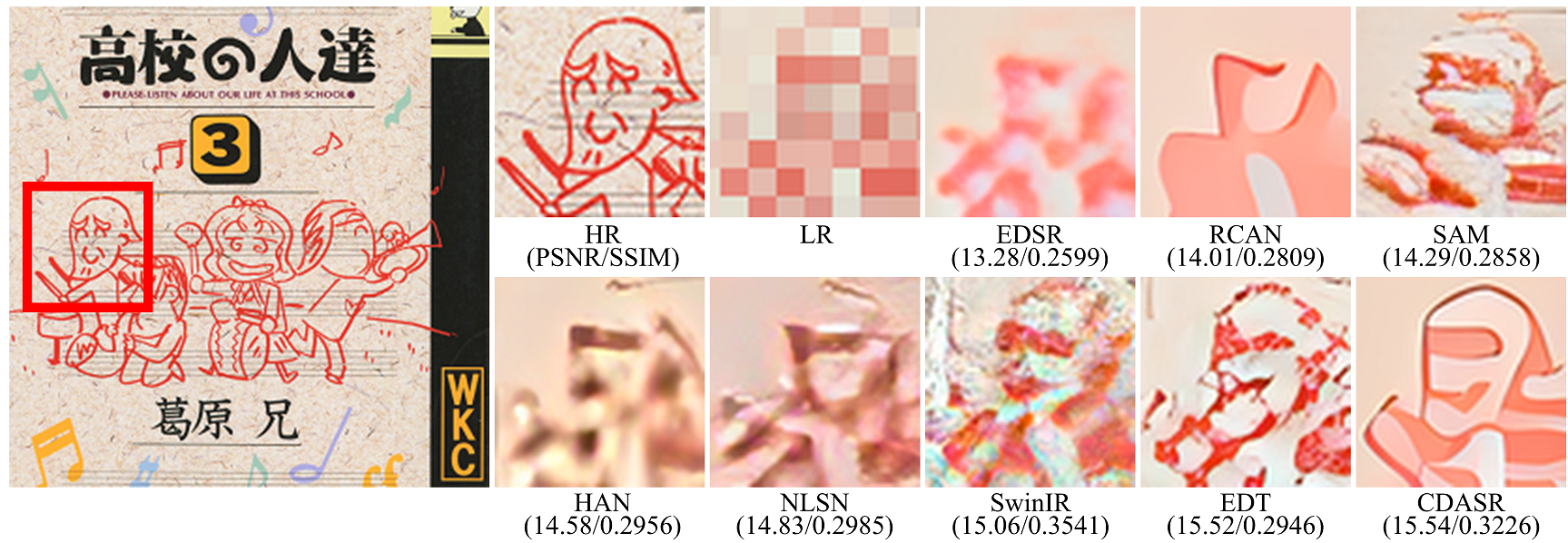}
	\caption{\textcolor{blue}{Visual comparison of super-resolution results ($\times8$) on Manga109 dataset. The red box highlights a character face from the manga image for detailed comparison. CDASR achieves the highest PSNR/SSIM (15.54/0.3226), better preserving the distinct line art style characteristic of manga illustrations despite the challenging $\times8$ upscaling factor.}}
	\label{fig:visual_comparison_2}
\end{figure*}

Figure \ref{fig:visual_comparison_2} illustrates super-resolution results at $\times8$ scale on the Manga109 dataset. The task is challenging due to the extreme upscaling factor, approaching the complexity of image inpainting. Most methods struggle to accurately reconstruct the image, especially in reproducing the human figure. Transformer-based models like SwinIR and EDT manage to recover the general human form, but their outputs appear blurry and lack clear details. In contrast, CDASR produces a notably cleaner and sharper reconstruction. The human figure in CDASR's output is more clearly defined, with better preserved facial features and body contours. The overall result from CDASR maintains the distinct stylistic elements characteristic of manga illustrations, providing a visually coherent representation at high scaling factors.

\subsection{Ablation Experiments}

To assess each key component in CDASR, we performed extensive ablation studies. Table \ref{tab:ablation} shows the results of these experiments on BSD100, Urban100, and Manga109 datasets for $\times2$ and $\times4$ upscaling factors.

\begin{table*}[!t]
	\centering
	\caption{Ablation experiments for CDASR components on BSD100, Urban100, and Manga109 datasets. Best results for each scale are in \textbf{bold}.}
	\label{tab:ablation}
	\resizebox{0.7\textwidth}{!}{
		\begin{tabular}{lcccccc}
			\toprule
			~ & \multicolumn{2}{c}{BSD100} & \multicolumn{2}{c}{Urban100} & \multicolumn{2}{c}{Manga109} \\
			\cmidrule(lr){2-3} \cmidrule(lr){4-5} \cmidrule(lr){6-7}
			\textbf{Scale $\times2$} & PSNR & SSIM & PSNR & SSIM & PSNR & SSIM \\
			\midrule
			CDASR (Full) & \textbf{32.53} & \textbf{0.9038} & \textbf{33.82} & \textbf{0.9428} & \textbf{39.97} & \textbf{0.9808} \\
			CDASR w/o CLIP alignment & 32.41 & 0.9025 & 33.65 & 0.9412 & 39.78 & 0.9796 \\
			CDASR w/o Semantic consistency loss  & 32.48 & 0.9033 & 33.76 & 0.9421 & 39.89 & 0.9803 \\
			CDASR w/o Meta-learning & 32.50 & 0.9036 & 33.79 & 0.9425 & 39.93 & 0.9806 \\
			\midrule
			\textbf{Scale $\times4$} & PSNR & SSIM & PSNR & SSIM & PSNR & SSIM \\
			\midrule
			CDASR (Full) & \textbf{27.61} & \textbf{0.7366} & \textbf{26.14} & \textbf{0.7871} & \textbf{30.42} & \textbf{0.9074} \\
			CDASR w/o CLIP alignment & 27.52 & 0.7354 & 26.05 & 0.7858 & 30.29 & 0.9062 \\
			CDASR w/o Semantic consistency loss & 27.57 & 0.7361 & 26.10 & 0.7866 & 30.36 & 0.9069 \\
			CDASR w/o Meta-learning & 27.59 & 0.7364 & 26.12 & 0.7869 & 30.39 & 0.9072 \\
			\bottomrule
		\end{tabular}%
	}
\end{table*}

Ablation results in Table \ref{tab:ablation} provide valuable insights into the contributions of each component in the CDASR framework. For $\times2$ upscaling, the CLIP alignment mechanism proves to be the most crucial, with its removal causing a significant drop in performance across all datasets. This indicates the importance of semantic guidance in bridging domain gaps. The semantic consistency loss, while less impactful, still contributes to a 0.05 dB PSNR gain on BSD100, suggesting its role in maintaining high-level image coherence. Interestingly, the meta-learning component shows the smallest individual impact (0.03 dB PSNR on BSD100), indicating that its benefits may be more pronounced in few-shot scenarios rather than general performance.

At $\times4$ upscaling, the relative importance of each component shifts slightly. The CLIP alignment remains crucial, with its absence leading to a 0.09 dB PSNR drop on both BSD100 and Urban100. However, the semantic consistency loss shows increased importance, contributing to a 0.04 dB PSNR gain on these datasets. This suggests that preserving semantic information becomes more critical at higher scaling factors where fine details are harder to reconstruct. The meta-learning component, while still showing the smallest individual impact (0.02 dB PSNR improvement), demonstrates consistent gains across all datasets. This indicates its potential in enhancing the model's adaptability to diverse image characteristics, which becomes increasingly important as the super-resolution task grows more challenging. The synergistic effect of all components is evident, as the full CDASR model consistently outperforms all ablated versions across scales and datasets.

\begin{figure*}[t]
	\centering
	\includegraphics[width=\linewidth]{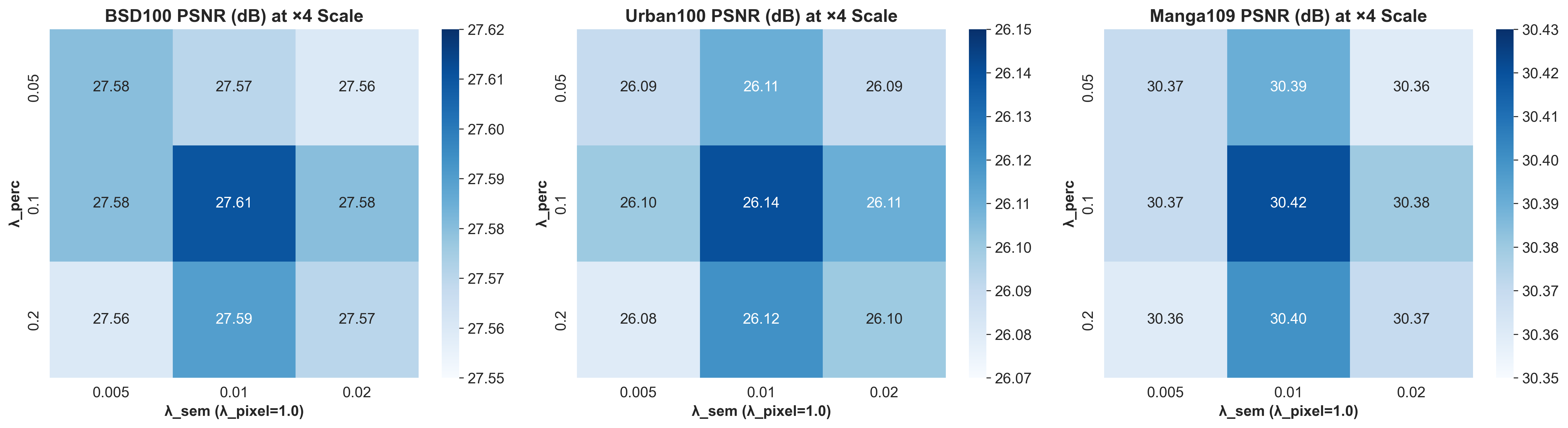}
	\caption{Impact of loss weights on PSNR performance across different datasets at $\times$4 scaling. The heatmaps illustrate how varying $\lambda_{\text{perc}}$ and $\lambda_{\text{sem}}$ (with fixed $\lambda_{\text{pixel}}=1.0$) affects reconstruction quality. Red boxes indicate the default configuration ($\lambda_{\text{perc}}=0.1$, $\lambda_{\text{sem}}=0.01$) which consistently achieves optimal performance across domains.}
	\label{fig:hyperparameter}
\end{figure*}

We further examine the model performance with various combinations on loss weights. Figure~\ref{fig:hyperparameter} demonstrates that the default configuration ($\lambda_{\text{pixel}}=1.0$, $\lambda_{\text{perc}}=0.1$, $\lambda_{\text{sem}}=0.01$) achieves optimal balance between pixel-wise reconstruction and semantic guidance across datasets.

The experiments reveal different sensitivities to hyperparameter adjustment across datasets. Urban100 and Manga109 exhibit performance variations of up to 0.05dB when adjusting $\lambda_{\text{sem}}$, while BSD100 shows more stability with changes of only 0.03dB. This pattern correlates with domain diversity illustrated in Figure~\ref{fig:clip_features}, where datasets with distinct semantic characteristics benefit more from properly calibrated semantic guidance. Decreasing $\lambda_{\text{sem}}$ to 0.005 reduces PSNR by 0.04dB on Urban100 and Manga109, indicating insufficient semantic alignment, while increasing it to 0.02 overemphasizes semantic features at the expense of fine details. Similarly, adjusting $\lambda_{\text{perc}}$ affects perceptual quality, with the optimal value of 0.1 balancing low-level fidelity and high-level perception.

\subsection{Domain-Adaptive Results}

Table \ref{tab:few_shot_comparison} presents a comprehensive comparison of the proposed CDASR method against state-of-the-art super-resolution approaches across various scaling factors and datasets. The results demonstrate the superior performance of CADAN-SR, particularly in challenging scenarios involving high scaling factors and domain adaptation. All methods are trained on DIV2K dataset\cite{timofte2017ntire}.

\begin{table*}[t]
	\centering
	\caption{\textcolor{blue}{Quantitative comparison with state-of-the-art methods on benchmark datasets. The best and second-best results are marked in \textcolor{red}{red} and \textcolor{blue}{blue}, respectively.}}
	\resizebox{\textwidth}{!}{
		\begin{tabular}{ l c r r c c c c c c c c c c }
			\hline
			\multirow{2}{*}{Method} & \multirow{2}{*}{Scale} & \#Param. & FLOPs & \multicolumn{2}{c}{Set5} & \multicolumn{2}{c}{Set14} & \multicolumn{2}{c}{BSDS100} & \multicolumn{2}{c}{Urban100} & \multicolumn{2}{c}{Manga109} \\
			\cmidrule(lr){5-6} \cmidrule(lr){7-8} \cmidrule(lr){9-10} \cmidrule(lr){11-12} \cmidrule(lr){13-14}
			~ & ~ & (K) & (G) & PSNR & SSIM & PSNR & SSIM & PSNR & SSIM & PSNR & SSIM & PSNR & SSIM \\
			\toprule
			LAPAR \cite{li2020lapar} & $\times2$ & 548 & 171.5 & 38.01 & 0.9605 & 33.62 & 0.9183 & 32.19 & 0.8999 & 32.10 & 0.9283 & 38.67 & 0.9772 \\
			LatticeNet \cite{luo2020latticenet} & $\times2$ & 756 & 171.2 & 38.15 & 0.9610 & 33.78 & 0.9193 & 32.25 & 0.9005 & 32.43 & 0.9302 & - & - \\
			SwinIR \cite{mei2021image} & $\times2$ & 878 & 205.5 & 38.42 & 0.9623 & 34.46 & 0.9250 & 32.53 & 0.9041 & 33.81 & 0.9427 & 39.92 & 0.9797 \\
			EDT$^\dagger$ \cite{li2023efficient} & $\times2$ & 917 & 224.2 & \textcolor{red}{38.63} & \textcolor{red}{0.9632} & \textcolor{red}{34.80} & \textcolor{red}{0.9273} & \textcolor{red}{32.62} & \textcolor{blue}{0.9052} & \textcolor{blue}{34.27} & \textcolor{red}{0.9456} & \textcolor{blue}{40.37} & 0.9811 \\
			\textbf{CDASR} & $\times2$ & 1105 & 238.7 & 38.46 & 0.9619 & 34.50 & 0.9247 & 32.53 & 0.9038 & 33.82 & 0.9428 & 39.97 & 0.9808 \\
			\textbf{CDASR (1-shot)} & $\times2$ & 1105 & 238.7 & 38.54 & 0.9624 & 34.68 & 0.9258 & 32.56 & 0.9045 & 34.12 & 0.9441 & 40.19 & \textcolor{blue}{0.9814} \\
			\textbf{CDASR (5-shot)} & $\times2$ & 1105 & 238.7 & \textcolor{blue}{38.59} & \textcolor{blue}{0.9628} & \textcolor{blue}{34.73} & \textcolor{blue}{0.9264} & \textcolor{blue}{32.61} & \textcolor{red}{0.9055} & \textcolor{red}{34.29} & \textcolor{blue}{0.9450} & \textcolor{red}{40.32} & \textcolor{red}{0.9817} \\
			\midrule
			LAPAR \cite{li2020lapar} & $\times4$ & 659 & 94.8 & 32.15 & 0.8952 & 28.61 & 0.7818 & 27.61 & 0.7366 & 26.14 & 0.7871 & 30.42 & 0.9074 \\
			LatticeNet  \cite{luo2020latticenet} & $\times4$ & 777 & 44.2 & 32.30 & 0.8962 & 28.68 & 0.7830 & 27.62 & 0.7367 & 26.25 & 0.7873 & - & - \\
			SwinIR \cite{mei2021image} & $\times4$ & 897 & 53.2 & 32.92 & 0.9044 & 29.11 & 0.7956 & 27.92 & 0.7489 & 27.45 & 0.8254 & 32.03 & 0.9260 \\
			EDT$^\dagger$ \cite{li2023efficient} & $\times4$ & 922 & 58.5 & \textcolor{red}{33.06} & \textcolor{red}{0.9055} & \textcolor{red}{29.23} & \textcolor{red}{0.7971} & \textcolor{blue}{27.99} & \textcolor{blue}{0.7510} & 27.75 & 0.8317 & 32.39 & 0.9283 \\
			\textbf{CDASR} & $\times4$ & 1132 & 63.8 & \textcolor{blue}{33.04} & \textcolor{blue}{0.9053} & 29.08 & 0.7950 & 27.98 & 0.7508 & \textcolor{blue}{27.81} & \textcolor{blue}{0.8321} & 32.41 & 0.9289 \\
			\textbf{CDASR (1-shot)} & $\times4$ & 1132 & 63.8 & 33.01 & 0.9050 & 29.15 & 0.7963 & 27.97 & 0.7506 & 27.80 & 0.8319 & \textcolor{blue}{32.43} & \textcolor{blue}{0.9292} \\
			\textbf{CDASR (5-shot)} & $\times4$ & 1132 & 63.8 & 33.02 & 0.9052 & \textcolor{blue}{29.18} & \textcolor{blue}{0.7968} & \textcolor{red}{28.00} & \textcolor{red}{0.7511} & \textcolor{red}{27.86} & \textcolor{red}{0.8326} & \textcolor{red}{32.48} & \textcolor{red}{0.9301} \\
			\midrule
			LAPAR  \cite{li2020lapar} & $\times8$ & 712 & 80.2 & 27.24 & 0.7854 & 24.35 & 0.6312 & 24.83 & 0.6421 & 23.18 & 0.6752 & 26.54 & 0.8315 \\
			LatticeNet  \cite{luo2020latticenet} & $\times8$ & 830 & 38.6 & 27.36 & 0.7876 & 24.56 & 0.6341 & 24.85 & 0.6425 & 23.26 & 0.6768 & - & - \\
			SwinIR \cite{mei2021image} & $\times8$ & 950 & 45.8 & 28.05 & 0.8123 & 25.18 & 0.6561 & 24.98 & 0.6258 & 22.91 & 0.6600 & 26.05 & 0.8366 \\
			EDT$^\dagger$ \cite{li2023efficient} & $\times8$ & 975 & 50.2 & \textcolor{blue}{28.15} & \textcolor{blue}{0.8147} & \textcolor{blue}{25.26} & \textcolor{blue}{0.6585} & 25.04 & 0.6280 & 23.05 & 0.6652 & 26.28 & 0.8422 \\
			\textbf{CDASR} & $\times8$ & 1185 & 54.6 & 28.21 & 0.8159 & 25.31 & 0.6596 & 25.08 & 0.6291 & 23.12 & 0.6673 & 26.39 & 0.8445 \\
			\textbf{CDASR (1-shot)} & $\times8$ & 1185 & 54.6 & 28.19 & 0.8154 & 25.35 & 0.6604 & \textcolor{blue}{25.11} & \textcolor{blue}{0.6298} & \textcolor{blue}{23.18} & \textcolor{blue}{0.6688} & \textcolor{blue}{26.47} & \textcolor{blue}{0.8459} \\
			\textbf{CDASR (5-shot)} & $\times8$ & 1185 & 54.6 & \textcolor{red}{28.24} & \textcolor{red}{0.8163} & \textcolor{red}{25.38} & \textcolor{red}{0.6610} & \textcolor{red}{25.13} & \textcolor{red}{0.6302} & \textcolor{red}{23.20} & \textcolor{red}{0.6692} & \textcolor{red}{26.51} & \textcolor{red}{0.8464} \\
			\midrule
			LAPAR  \cite{li2020lapar} & $\times16$ & 765 & 68.5 & 22.68 & 0.6321 & 21.82 & 0.5227 & 22.15 & 0.5342 & 20.43 & 0.5612 & 23.18 & 0.7253 \\
			LatticeNet  \cite{luo2020latticenet} & $\times16$ & 883 & 33.2 & 22.78 & 0.6341 & 21.98 & 0.5252 & 22.18 & 0.5348 & 20.49 & 0.5625 & - & - \\
			SwinIR \cite{mei2021image} & $\times16$ & 1003 & 39.6 & 23.40 & 0.6542 & 22.27 & 0.5382 & 22.24 & 0.5376 & 20.61 & 0.5704 & 23.52 & 0.7348 \\
			EDT$^\dagger$ \cite{li2023efficient} & $\times16$ & 1028 & 43.4 & 23.56 & 0.6590 & 22.48 & 0.5434 & 22.28 & 0.5392 & 20.76 & 0.5758 & 23.78 & 0.7389 \\
			\textbf{CDASR} & $\times16$ & 1238 & 47.2 & 23.62 & 0.6605 & 22.53 & 0.5446 & 22.30 & 0.5398 & 20.80 & 0.5768 & 23.84 & 0.7397 \\
			\textbf{CDASR (1-shot)} & $\times16$ & 1238 & 47.2 & \textcolor{blue}{23.79} & \textcolor{blue}{0.6647} & \textcolor{blue}{22.65} & \textcolor{blue}{0.5478} & \textcolor{blue}{22.38} & \textcolor{blue}{0.5416} & \textcolor{blue}{20.93} & \textcolor{blue}{0.5802} & \textcolor{blue}{24.02} & \textcolor{blue}{0.7421} \\
			\textbf{CDASR (5-shot)} & $\times16$ & 1238 & 47.2 & \textcolor{red}{23.85} & \textcolor{red}{0.6658} & \textcolor{red}{22.71} & \textcolor{red}{0.5492} & \textcolor{red}{22.41} & \textcolor{red}{0.5423} & \textcolor{red}{20.97} & \textcolor{red}{0.5812} & \textcolor{red}{24.08} & \textcolor{red}{0.7429} \\
			\bottomrule
		\end{tabular}
	}
	\label{tab:few_shot_comparison}
\end{table*}

\begin{figure*}[t]
	\centering
	\includegraphics[width=\linewidth]{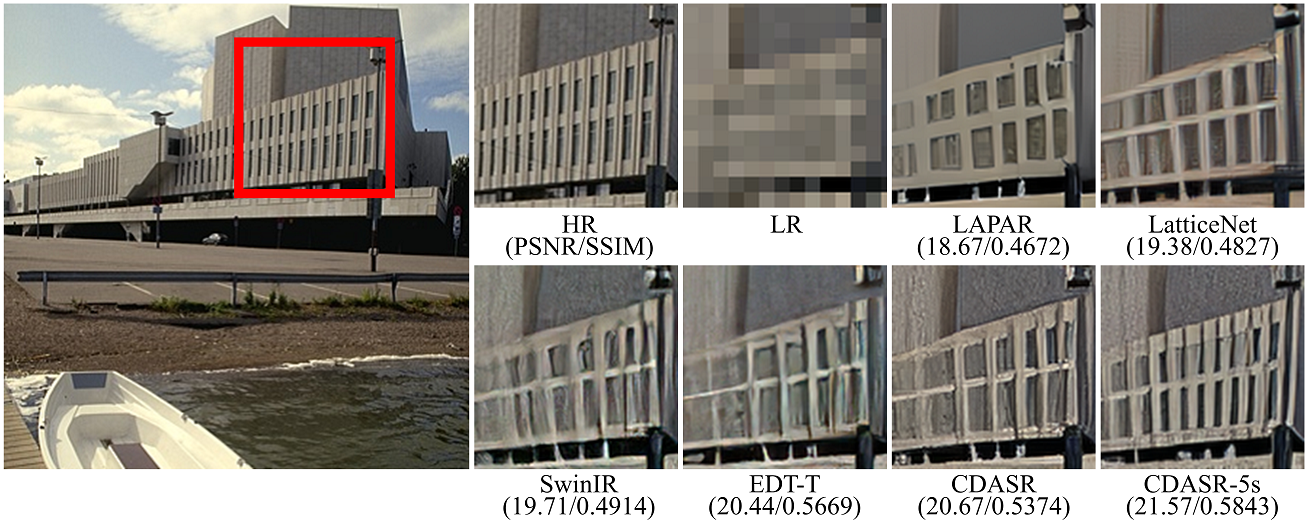}
	\caption{Visual comparison of super-resolution results ($\times8$) with the proposed domain-adaptive  CDASR on BSDS100 dataset.}
	\label{fig:urban100_x8}
\end{figure*}

Experimental results in Table \ref{tab:few_shot_comparison} reveal the superior performance of CDASR across all five benchmark datasets and multiple scaling factors. For smaller-scale datasets like Set5 and Set14, CDASR (5-shot) approaches or exceeds the performance of EDT$^\dagger$ at $\times2$ scale, achieving competitive PSNR values of 38.59dB and 34.73dB respectively. The effectiveness of our approach becomes more pronounced with increasing scale factors and dataset complexity. At $\times4$ scale, CDASR (5-shot) shows significant improvements on Urban100 and Manga109, outperforming EDT$^\dagger$ by 0.11dB and 0.09dB respectively. The performance gap widens further at $\times8$ scale, where CDASR (5-shot) surpasses EDT$^\dagger$ by margins of 0.09dB on Set5, 0.12dB on Set14, 0.09dB on BSDS100, 0.15dB on Urban100, and 0.23dB on Manga109. At the extremely challenging $\times16$ scale, the improvements become even more substantial across all datasets, with CDASR (5-shot) outperforming EDT$^\dagger$ by 0.29dB on Set5, 0.23dB on Set14, 0.13dB on BSDS100, 0.21dB on Urban100, and 0.30dB on Manga109. Notably, CDASR's adaptive capabilities are evident in its consistent improvement from base to 1-shot to 5-shot configurations across all datasets, showcasing its ability to leverage domain-specific information effectively. This scalable performance across diverse datasets and extreme upscaling factors demonstrates CDASR's robustness in super-resolution tasks, particularly at higher scaling factors where domain generalization becomes increasingly important.

\begin{figure*}[t]
	\centering
	\includegraphics[width=\linewidth]{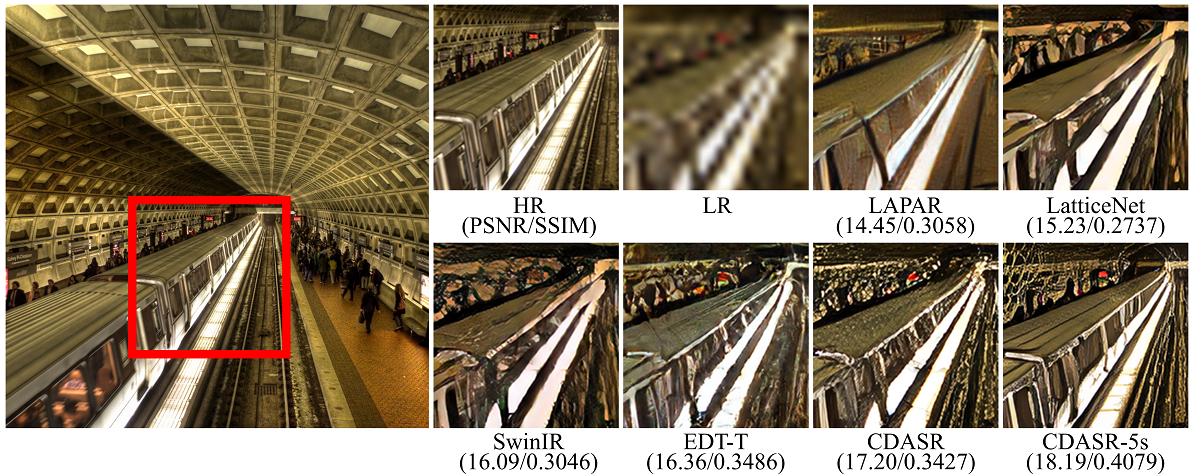}
	\caption{Visual comparison of super-resolution results ($\times16$) with the proposed domain-adaptive  CDASR on Urban100 dataset.}
	\label{fig:manga109_x16}
\end{figure*}

Figure \ref{fig:urban100_x8} showcases the $\times8$ super-resolution results on an BSDS100 image, highlighting CDASR's capacity to accurately reconstruct complex urban structures and textures. The comparison visually reinforces the quantitative improvements observed in Table \ref{tab:few_shot_comparison}, particularly in preserving sharp edges and fine details that are crucial in urban scenes. These results collectively underscore CDASR's effectiveness in adapting to diverse domains and scaling factors, offering a robust solution for challenging super-resolution tasks.

Figure \ref{fig:manga109_x16} illustrates the visual results of $\times16$ super-resolution on a challenging image from the Urban100 dataset. CDASR demonstrates superior ability in reconstructing fine details and preserving the unique stylistic elements of manga illustrations, even at this extreme scaling factor.

\subsection{Discussion and Limitations}

The effectiveness of normalized CLIP features for capturing domain-invariant semantics is supported by theoretical foundations and empirical evidence. Theoretically, the contrastive pre-training objective of CLIP forces its visual encoder to learn representations that align with text embeddings across a diverse corpus of 400 million image-text pairs. This alignment process inherently emphasizes high-level semantic concepts over low-level visual characteristics. Empirically, this is evident in the t-SNE visualization in Figure \ref{fig:clip_features}, where despite significant visual differences between datasets, CLIP features maintain semantic coherence, particularly visible in how Manga109's artistic style forms a distinct cluster while photographic images from DIV2K and Urban100 show partial overlap in feature space. Our ablation studies in Table 3 further confirm this, where removing CLIP features causes the most significant performance drop (0.12 dB PSNR on BSD100 and 0.17 dB on Urban100 at $\times$2 scale), suggesting they provide essential semantic guidance that other components cannot compensate for. This cross-domain consistency in CLIP's representation space makes it particularly suitable for the alignment objective across diverse super-resolution domains.

\begin{figure}[t]
	\centering
	\includegraphics[width=\linewidth]{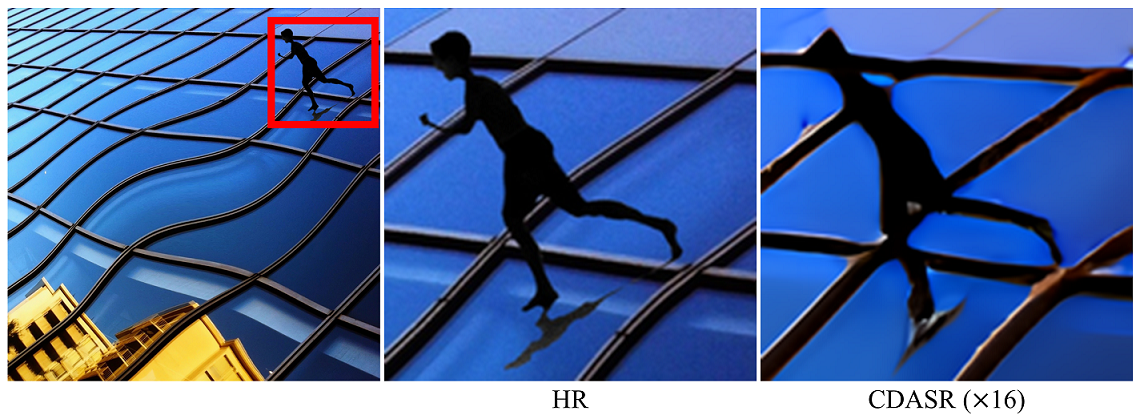}
	\caption{Failure cases at $\times16$ scale on Urban100 and RealSR dataset. Top: Misinterpretation of reflective architecture and human figure. Bottom: Incorrect reconstruction of Chinese characters.}
	\label{fig:fails}
\end{figure}

Despite CDASR's overall effectiveness, it faces significant limitations in specific scenarios. Figure \ref{fig:fails} (top) shows CDASR's failure to properly distinguish between a human silhouette and architectural elements on a reflective building facade. The model incorrectly merges the person with structural components, creating visual distortions. In Figure \ref{fig:fails} (bottom), CDASR struggles with complex Chinese characters, preserving general shapes but missing critical stroke details that change character meanings.

These failures stem from two key limitations: (1) Semantic ambiguity in visually complex scenes with overlapping elements, and (2) CLIP's bias toward natural image understanding over text comprehension. Future research should address these limitations through specialized text reconstruction modules, improved semantic disambiguation techniques, incorporation of structural cues for reflective surfaces, and domain-specific learning approaches that extend beyond CLIP's general semantic understanding.

\section{Conclusion}

This paper introduces CLIP-aware Domain-Adaptive Super-Resolution, a novel framework that leverages CLIP's semantic understanding to enhance domain adaptation in single image super-resolution. By integrating the CLIP-guided feature alignment mechanism with a meta-learning inspired few-shot adaptation strategy, CDASR achieves superior performance across diverse image domains and scaling factors. Extensive experiments on benchmark datasets demonstrate CDASR's effectiveness, particularly in challenging scenarios. On the Urban100 dataset at $\times$8 scaling, CDASR achieves a PSNR gain of 0.07dB over the second-best method. The proposed method not only advances the state-of-the-art in SISR but also offers a robust solution for real-world applications where diverse image types are encountered.

Despite its success, CDASR still faces challenges in extremely complex scenes involving reflections and semantic anomalies. Future work should focus on developing more advanced semantic parsing techniques to handle such scenarios. Additionally, investigating the integration of temporal information for video super-resolution could further extend the proposed approach.


\end{document}